\documentclass[runningheads]{llncs}
\usepackage[T1]{fontenc}
\usepackage{cite}
\usepackage{amsmath,amssymb,amsfonts}
\usepackage{algorithmic}
\usepackage{algorithm}
\usepackage{textcomp}
\usepackage{hyperref}
\usepackage{pifont}
\usepackage{amsmath,amssymb} 
\usepackage{physics} 
\usepackage{xspace} %
\usepackage{bm} 
\usepackage{graphicx,verbatim}
\usepackage{caption}
\usepackage{array}

\usepackage[table,xcdraw,dvipsnames]{xcolor}
\usepackage{colortbl} 

\begin{document}
\title{Hallucination-Aware Multimodal Benchmark for Gastrointestinal Image Analysis with Large Vision-Language Models}

\titlerunning{Hallucination-Aware Multimodal Benchmark for GI Analysis}

\author{Bidur Khanal\inst{1}\textsuperscript{*} \and 
Sandesh Pokhrel\inst{2}\textsuperscript{*} \and 
Sanjay Bhandari\inst{2}\textsuperscript{*} \and 
Ramesh Rana\inst{3} \and 
Nikesh Shrestha\inst{3} \and 
Ram Bahadur Gurung\inst{3} \and 
Cristian Linte\inst{1} \and 
Angus Watson\inst{5} \and 
Yash Raj Shrestha\inst{4} \and 
Binod Bhattarai\inst{5}}

\authorrunning{Khanal, Pokhrel, Bhandari et al.}

\institute{Rochester Institute of Technology, Rochester, NY, USA \and 
Nepal Applied Mathematics and Informatics Institute for Research (NAAMII), Lalitpur, Nepal \and 
Kathmandu University, Dhulikhel, Nepal \and 
University of Lausanne, Lausanne, Switzerland \and 
University of Aberdeen, Aberdeen, UK}

\renewcommand{\thefootnote}{\fnsymbol{footnote}}
\footnotetext[1]{These authors contributed equally to this work. Khanal was associated with Multimodal Learning Lab, UoA, while working on this paper. Corresponding author email: \email{binod.bhattarai@abdn.ac.uk}}

\maketitle              
\begin{abstract}
Vision-Language Models (VLMs) are becoming increasingly popular in the medical domain, bridging the gap between medical images and clinical language. Existing VLMs demonstrate an impressive ability to comprehend medical images and text queries to generate detailed, descriptive diagnostic medical reports. However, hallucination--the tendency to generate descriptions that are inconsistent with the visual content--remains a significant issue in VLMs, with particularly severe implications in the medical field. To facilitate VLM research on gastrointestinal (GI) image analysis and study hallucination, we curate a multimodal image-text GI dataset: Gut-VLM. This dataset is created using a two-stage pipeline: first, descriptive medical reports of Kvasir-v2 images are generated using ChatGPT, which introduces some hallucinated or incorrect texts. In the second stage, medical experts systematically review these reports, and identify and correct potential inaccuracies to ensure high-quality, clinically reliable annotations. Unlike traditional datasets that contain only descriptive texts, our dataset also features tags identifying hallucinated sentences and their corresponding corrections. A common approach to reducing hallucination in VLM is to finetune the model on a small-scale, problem-specific dataset. However, we take a different strategy using our dataset. Instead of finetuning the VLM solely for generating textual reports, we finetune it to detect and correct hallucinations, an approach we call hallucination-aware finetuning. Our results show that this approach is better than simply finetuning for descriptive report generation. Additionally, we conduct an extensive evaluation of state-of-the-art VLMs across several metrics, establishing a benchmark. GitHub Repo: \href{https://github.com/bhattarailab/Hallucination-Aware-VLM}{bhattarailab/Hallucination-Aware-VLM.}

\keywords{Multimodal data  \and Gastrointestinal image analysis \and Vision-Language Model (VLM) \and Hallucination \and Hallucination-aware finetuning}

\end{abstract}
\section{Introduction}
\label{sec:introduction}

Gastrointestinal (GI) diseases affect millions of people globally, making an accurate and timely diagnosis crucial for effective patient care \cite{arnold2020global}. GI endoscopy is the gold standard tool for diagnosing gastrointestinal diseases and is widely adopted in clinical settings. In recent years, Artificial Intelligence (AI) has shown significant potential in assisting clinicians with disease understanding and decision-making by detecting conditions \cite{gitractensemblestacking, svmcnngastrodisease}, classifying anatomical landmarks \cite{giclassification}, and identifying anomalies \cite{AnamolyasObjectdet,selfsupervisedood,ttaood,ncdd}.  While AI models primarily rely on endoscopic images, integrating descriptive text enhances expressiveness and interpretability, providing a richer clinical context that enables informed clinical decision making~\cite{li2022deep}, diagnostic support~\cite{marques2017image,lorraine2023diagnostic}, quality assurance~\cite{marques2017image}, communication~\cite{selivanov2023medical}, medical record documentation~\cite{marques2017image}, and more. Nevertheless, despite the importance of textual information, its practical usage remains limited due to the lack of such multimodal datasets containing both GI endoscopic images and descriptive texts.

Several medical image-text datasets exist for chest X-rays images, histopathology images, and other radiographs \cite{MeidcalVLpretraining}, enabling the development of VLMs for clinical applications \cite{singhal2025toward, zebra_healthcare}. Unlike other image-only AI tools, VLMs are inherently good at absorbing complex information, reasoning, and generating explanations that are comprehensive for both clinicians and patients. Although several GI image and video datasets exist \cite{hyperkvasir,kvasir-capsule,kvasir,kvasir-instrument,gastrovision,multimodalimage}, to the best of our knowledge, Kvasir-VQA \cite{kvasirvqa} is the only existing text-image multimodal dataset for GI image analysis, but it has notable limitations. Its short textual responses limit the depth of expert analysis and do not fully accommodate specialized medical vocabulary. Moreover, this dataset lacks comprehensive validation of all samples by certified experts, attributed to time constraints. 

To mitigate these shortcomings, we create a new multimodal dataset out of Kvasir-v2 images, describing the underlying conditions in a short descriptive report format. Instead of having experts manually annotate images from the start, we first leverage an existing commercial large VLM (ChatGPT-4 Omni \cite{achiam2023gpt}) to generate image descriptions by prompting it with expert-crafted questions, which serves as a cost-effective and time-efficient alternative. Studies show that ChatGPT, trained on vast internet data, demonstrates a reasonable understanding of medical prompts and performs decently on some evaluations \cite{shieh2024assessing,keshavarz2024chatgpt}. However, like any other language models, VLMs are prone to hallucinations and cannot be relied upon without certified expert supervision.

Hallucination, in this context, refers to instances where the model produces information that appears plausible but is factually incorrect or fabricated \cite{gunjal2024detecting,hallucinationmultimodallargelanguage}. VLMs generate outputs in an auto-regressive manner, predicting the next plausible token based on statistical patterns. However, this process can introduce biases, as predictions tend to favor patterns that are most frequently observed in the training data rather than being grounded in factual knowledge. A recent study indicates that the state-of-the-art (SOTA) large VLMs exhibit a hallucination rate up to 30\% when describing natural images~\cite{gunjal2024detecting}. Hallucination can be more severe in medical data, as demonstrated by our dataset (Fig. \ref{fig:pipeline}), where only $30.39\%$ of the VLM-generated responses are fully correct. To address this issue, in the second stage, we employ expert gastroenterologists to analyze the images and responses generated by ChatGPT, identify potential hallucinations, and correct any inaccuracies, ensuring that the final response is accurate and reliable. As a result, we obtain expert feedback on where the VLM hallucination has occurred at the sentence level and what the corrected version is. This additional information in the dataset can be leveraged to develop hallucination detectors or create hallucination-aware models.

Some studies in educational psychology suggest that students learn effectively when actively engaged in correcting errors and self-reflecting rather than through passive learning \cite{metcalfe2017learning}. We argue that training VLM to identify and correct hallucinations fosters learning through correction, similar to how humans learn. Our hallucination-aware strategy for finetuning VLMs is rooted in this motivation.  

While several datasets and studies on hallucination in natural image-text scenarios have recently emerged \cite{hallucinationmultimodallargelanguage}, there have been very few attempts to explore this phenomenon within the medical domain \cite{hallucinationinmedical,comthallucination}, and none specifically in GI analysis. 
Therefore, in this work, we propose a hallucination-aware multimodal dataset, Gut-VLM, for GI image analysis using Kvasir-v2 images, one of the most widely used GI image datasets, to facilitate research on the clinical applications of VLMs and study hallucinations. Our key contributions are:

\begin{enumerate} \item We create a novel multimodal image-text dataset for GI image analysis, consisting of VLM-generated descriptive diagnostic reports, expert-labeled tags identifying hallucinated sentences, and their corresponding corrections. 

\item We provide an extensive evaluation benchmark on four SOTA VLMs across various settings, using both existing and our proposed LLM-assisted evaluation metrics, alongside a clinical expert evaluation.

\item We demonstrate that our innovative hallucination-aware finetuning approach, trained to detect and correct hallucinations, improves test performance compared to finetuning only on corrected ground-truth responses. \end{enumerate}

\section{Related Works}

Several GI image and video datasets, such as HyperKvasir \cite{hyperkvasir}, Kvasir-Capsule \cite{kvasir-capsule}, Kvasir \cite{kvasir}, Kvasir-Instrument \cite{kvasir-instrument}, GastroVision \cite{gastrovision}, and EndoAngel-MM \cite{multimodalimage}, have facilitated research in AI-assisted diagnosis and surgical intervention for GI endoscopy. The recent shift toward using VLMs for clinical analysis of GI images through textual descriptions is gradually gaining traction \cite{kvasirvqa}, despite remaining largely unexplored. As the use of VLMs becomes more prominent in the medical domain, challenges such as hallucination and trustworthiness are emerging as important issues. 

Recently, many datasets and benchmarks have been created to study hallucinations in VLM in natural image settings\cite{gunjal2024detecting}. Works also include techniques to address hallucinations, such as incorporating negative examples in training \cite{LRVInstruction}, reducing noise \cite{eos,Recaption}, and using counterfactual data \cite{hallucidoctor}, better feature integration \cite{mmllversatile}, architectural change \cite{llava1.5}, and others. The evaluation benchmarks typically used classical metrics like CHAIR \cite{CHAIR} and POPE \cite{POPE}, which rely on the COCO dataset \cite{lin2014microsoft}. More recent works have also introduced LLM-assisted evaluation techniques, including CIEM \cite{CIEM}, GAIVE \cite{Gaive}, and FaithScore \cite{faithscore}. Primarily designed for evaluating generic textual responses, these metrics are not readily applicable to the medical domain. Recently, Chen et al. \cite{hallucinationinmedical} proposed a benchmark for general medical hallucination in VLM, while Jiang et al. \cite{comthallucination} demonstrated the effectiveness of chain-of-thought training in mitigating hallucinations during medical report generation. However, these benchmarks do not cover GI images and rely on non-standardized, generic evaluation frameworks to categorize hallucination types, making them unsuitable for GI image analysis.
 
\section{Methodology}

In this section, we introduce the Gut-VLM dataset, detailing its overall composition and outlining the annotation pipeline, and present some key dataset statistics, particularly focusing on VLM-induced hallucinations. Finally, we also present a hallucination-aware VLM finetuning strategy.

\begin{figure}[ht!]
    \centering
    \includegraphics[width=\linewidth]{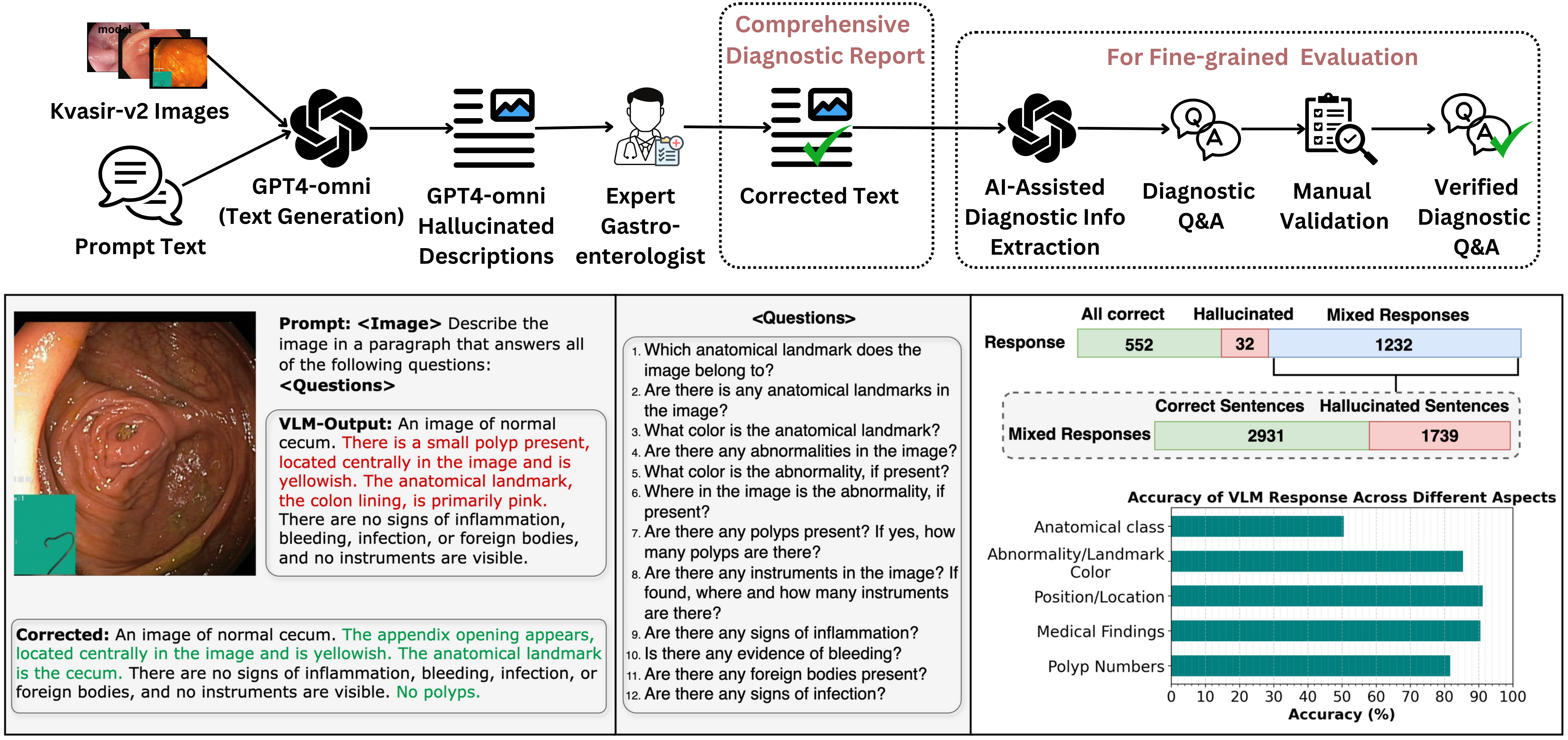}
    \caption{Top: Overview of the data annotation pipeline. Bottom: [Left] A sample from the Gut-VLM, describing the underlying conditions based on the questions asked in the prompt, showing hallucinated (\textcolor{BrickRed}{red}) and corrected texts (\textcolor{ForestGreen}{green}). [Right] Statistics of ChatGPT-4 Omni-generated responses in the dataset.}
    \label{fig:pipeline}
    \vspace{-1em}
\end{figure}

\subsection{Dataset Composition}

The images in the Gut-VLM dataset are sourced from Kvasir-v2 \cite{kvasir}, covering a diverse set of normal and abnormal gastrointestinal conditions. Normal findings include images of healthy Cecum, Pylorus, and Z-line anatomy, while abnormal findings include images of Polyps, Esophagitis, Ulcerative Colitis, Dyed-resection-margins, and Dyed-lifted-polyps. We annotated a total of 1,816 images representing these conditions, splitting them into 1,450 for training and 366 for testing. Instead of randomly splitting the entire dataset into training and test sets, we ensured proportional representation of each condition in the test set by allocating $20\%$ of samples from each category to the test set.

\subsection{Dataset Annotation Pipeline}
In this section, we outline the multistep pipeline used to generate the final annotated dataset, as illustrated in Fig. \ref{fig:pipeline}. This innovative, cost-effective approach involves using a large VLM to generate descriptive diagnostic reports for images, followed by expert corrections to produce verified reports. The report is further parsed to extract diagnostic Q\&A for fine-grained evaluation.\\

\noindent\textbf{VLM Diagnostic Report Generation:} \label{vlm_report} To generate a descriptive report for each image in our dataset, we queried GPT-4 Omni \cite{achiam2023gpt} to describe each image using a prompt designed to elicit a detailed response covering the contents of 12 diagnostic reference questions for gastrointestinal image assessment. These questions, part of the MedVQA-GI Challenge \cite{Medvqa}, address aspects such as anatomical class, color and position of landmarks and abnormalities, polyp count, and the presence of inflammation, bleeding, foreign bodies, infection, or instruments. Since the generated responses contained some hallucinated texts, we proceeded to the next step to identify and correct these hallucinations.

The VLM-generated responses are reviewed by expert gastroenterologists to identify hallucinations. Following the M-HalDetect framework \cite{gunjal2024detecting}, each sentence in the response was labeled to indicate whether it contained hallucinated text. A sentence is marked as non-hallucinated if it accurately describes the content in the image, while any sentence containing inconsistent information is marked as hallucinated. Fig. \ref{fig:pipeline} presents the corresponding hallucination statistics for the overall dataset: only $30.39\%$ of the responses were fully correct, $1.7\%$ contained hallucinations in all sentences, and $67.84\%$ consisted of mixed responses with correct and hallucinated sentences. Finally, experts corrected incorrect sentences to ensure that the descriptions are clinically accurate. These annotations were collected using our in-house developed annotation tool and involved four expert gastroenterologists.

\noindent\textbf{Diagnostic Q\&A Extraction for Fine-Grained Evaluation:} \label{qna_extract}While diagnostic reports are more comprehensive and context-rich, to objectively evaluate performance at a fine-grained level, we must assess whether the descriptions accurately address the 12 diagnostic questions. To achieve this, we prompted ChatGPT to extract information from the generated descriptive responses into a structured Q\&A format. This method was also applied to extract diagnostic Q\&A for other VLMs during testing. For the corrected ground-truth responses, we manually verified the extracted diagnostic Q\&A to ensure accuracy. We observed a hallucination rate of $4.29\%$ during the information extraction process from the corresponding descriptive responses. As the same extractor was applied to all models during evaluation, the hallucination impact should be consistent across models.  Additionally, since the results are available in a structured Q\&A format, this data can also be utilized for Visual Question Answering (VQA) experiments.

\begin{figure}[h!]
    \centering
    \includegraphics[width=0.9\linewidth]{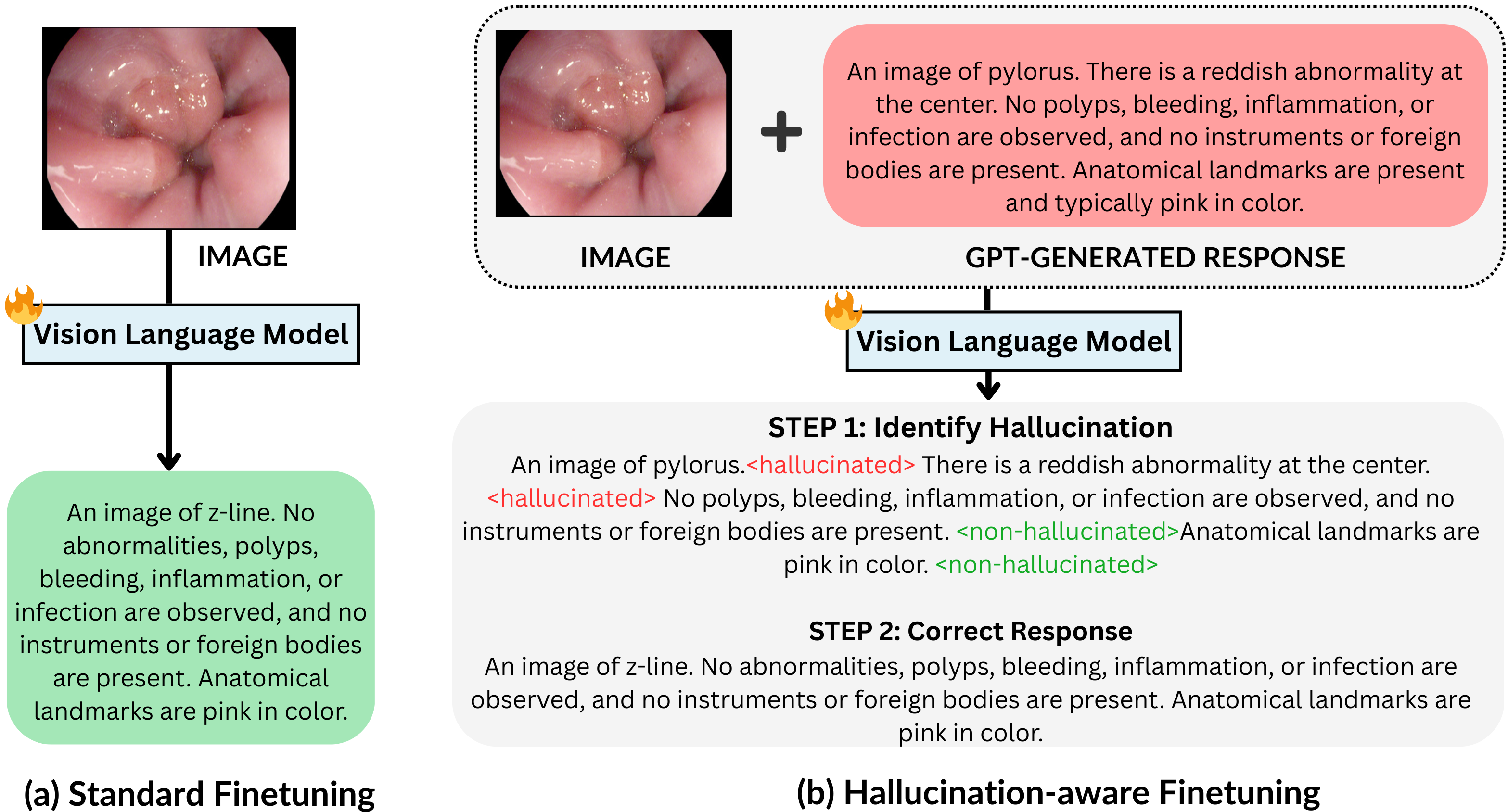}
    \caption{Comparing standard finetuning with hallucination-aware finetuning}
    \label{fig:hal-aware}
    \vspace{-1em}
\end{figure}

\subsection{Hallucination-aware VLM Finetuning:}

A standard VLM finetuning approach for generating descriptions would be to train the model to output the ground-truth texts. However, since we have the VLM-generated response, sentence-level hallucination tags, and the final corrected response, we instead finetune the model to identify hallucinated sentences and correct the response, as shown in Fig. \ref{fig:hal-aware}. This approach leverages the existing VLM responses and makes the model aware of potential hallucination patterns.

\section{Experiments}

Using our proposed dataset, we experimented with four state-of-the-art (SOTA) VLM models: LLaVA-1.6-7B \cite{liu2023visual}, Qwen2-7B \cite{yang2024qwen2technicalreport}, mPLUG-Owl-2B \cite{ye2024mplugowl3longimagesequenceunderstanding}, and DeepSeek-7B-VL \cite{deepseekai2025deepseekv3technicalreport}. We first generated descriptive reports by prompting the pretrained models to describe images with a focus on 12 diagnostic questions. Next, we applied supervised finetuning using Rank-8 LoRA adaptation \cite{hu2022lora} with two strategies. In the first strategy, we finetuned the model to generate corrected responses directly. In the second strategy, we finetuned the model to learn to detect hallucinated sentences and then correct them—the approach we refer to as hallucination-aware (H) finetuning.

As outlined in Section ~\ref{qna_extract}, since the dataset was already structured in a VQA format for fine-grained evaluation, we explicitly experimented with one of the models, LLaVA-1.6-7B, on a VQA task. We first tested the pretrained model, then applied standard VQA finetuning in a multi-conversational format.

All models were finetuned using an A100 GPU for five epochs until the loss stabilized, with a batch size of 8 and a learning rate of 1e-4, and evaluated on the test set. We used the ms-swift framework \footnote{https://github.com/modelscope/ms-swift} to run all VLM experiments.

\section{Evaluation}

We initially assess VLM-generated reports against corrected ground-truth descriptions using classical metrics such as ROUGE-L \cite{lin2004rouge}, BLEU\cite{papineni2002bleu}, and METEOR\cite{banerjee2005meteor}, which measure sequence overlap, n-gram matches, and semantic similarity, respectively. However, these metrics are limited by context-length dependence, insensitivity to subtle semantic differences, and inability to assess factual accuracy. Here, we propose two \textcolor{blue}{LLM-assisted evaluation metrics}. The first, \textcolor{red}{R-Sim}, rates coarse-level semantic similarity between the ground-truth and VLM-generated responses on a scale of 1 to 5 (worst to best), using ChatGPT-4 prompted to assess similarity with a focus on GI endoscopy and 12 diagnostic questions. Such LLM-assisted evaluations have been reported in recent studies\cite{Gaive}. The second, \textcolor{red}{Question Answering Accuracy Score (QAAS)}, objectively measures accuracy by comparing VLM responses to 12 ground-truth Q\&A pairs, with ChatGPT handling synonyms and similar phrasing. Any descriptions are first converted and parsed to Q\&A format, following Section~\ref{qna_extract}. 

\textit{We had an expert rate LLaVa's descriptive responses for clinical evaluation. Due to budget constraints, we randomly sampled 30 responses per model and asked the expert to provide two ratings (1 to 5): \textcolor{red}{similarity}, assessing the clinical resemblance of the response to the ground truth, and \textcolor{red}{quality}, evaluating its clinical significance independently. We averaged the two ratings to compute a final per-response score. Additionally, a new sample was randomly inserted among the 30 responses five times, unknown to the expert, to measure rating consistency.}

\section{Results}

\subsubsection{Descriptive Format Diagnostic Report:}
Table \ref{tab:Results} presents the quantitative evaluation of descriptive reports generated by VLMs for GI images. All pretrained models exhibited significantly lower performance in terms of ROUGE-L, BLEU, METEOR, R-Sim, and QAAS scores, demonstrating a limited ability to generate clinically relevant descriptions—except for ChatGPT-4 Omni. Since ChatGPT-4 is a large-parameter model and the ground-truth descriptions are corrected versions of the original ChatGPT responses, higher scores are expected due to potential bias toward the original structure.

\begin{table}[h!]
    \vspace{-2em}
    \centering
    \setlength{\tabcolsep}{1pt} 
    \caption{Quantitative evaluation of various VLMs in descriptive report generation and VQA tasks. Finetuned$^\textbf{H}$ indicates hallucination-aware finetuning.}
    \begin{tabular}{l@{\hspace{-9pt}}ccc>{\columncolor[gray]{0.9}}c>{\columncolor[gray]{0.9}}c}
    
    \hline 
        \textbf{Model} & \textbf{ROUGE-L$\uparrow$} & \textbf{BLEU$\uparrow$} & \textbf{METEOR$\uparrow$} & \textbf{R-Sim$\uparrow$} & \textbf{QAAS($\%$)$\uparrow$} \\ \hline
         ChatGPT-4 \textit{Omni} & $0.87$ & $0.80$ & $0.89$ & $2.97$ & $85.99$\\ \hline
         LLaVa-1.6-7b \textit{pretrained} & $0.26$ & $0.10$ & $0.47$ & $1.36$ & $50.89$ \\
         LLaVa-1.6-7b \textit{finetuned} & $0.54$ & $0.35$ & $0.63$ & $3.71$ & $83.07$ \\
         LLaVa-1.6-7b \textit{finetuned}$^\textbf{H}$ & $\mathbf{0.89}$ & $\mathbf{0.82}$ & $\mathbf{0.90}$ & $\mathbf{3.96}$ & $\mathbf{90.89}$  \\
         \hline
         DeepSeek7b \textit{pretrained} & $0.29$ & $0.11$ & $0.39$ & $1.65$ & $65.20$ \\
         DeepSeek7b \textit{finetuned} & $0.55$ & $0.37$ & $0.65$ & $\mathbf{3.76}$ & $83.73$ \\
         DeepSeek7b \textit{finetuned}$^\textbf{H}$ & $\mathbf{0.88}$ & $\mathbf{0.81}$ & $\mathbf{0.90}$ & $3.63$ & $\mathbf{88.77}$ \\
         \hline
         Qwen7b \textit{pretrained} & $0.32$ & $0.12$ & $0.48$ & $1.74$ & $67.57$ \\
         Qwen7b \textit{finetuned} & $0.54$ & $0.37$ & $0.64$ & $3.78$ & $83.27$ \\
         Qwen7b \textit{finetuned}$^\textbf{H}$ & $\mathbf{0.88}$ & $\mathbf{0.82}$ & $\mathbf{0.90}$ & $\mathbf{4.04}$ & $\mathbf{90.53}$ \\
         \hline
         MPlugOwl2b \textit{pretrained} & $0.26$ & $0.09$ & $0.44$ & $1.34$ & $55.29$ \\
         MPlugOwl2b \textit{finetuned} & $0.50$ & $0.32$ & $0.60$ & $3.68$ & $82.90$ \\
         MPlugOwl2b \textit{finetuned}$^\textbf{H}$ & $\mathbf{0.85}$ & $\mathbf{0.77}$ & $\mathbf{0.87}$ & $\mathbf{3.72}$ & $\mathbf{88.40}$  \\         
         \hline
         VQA LLaVa-1.6-7b \textit{pretrained} & $-$ & $-$ & $-$ & $-$ & $49.26$   \\
         VQA LLaVa-1.6-7b \textit{finetuned} & $-$ & $-$ & $-$ & $-$ & $87.91$  \\
         \hline
        
    \end{tabular}
    \label{tab:Results}
    \vspace{-1em}
\end{table}

\begin{figure}[h!]
    \centering
    \includegraphics[width=0.95\linewidth]{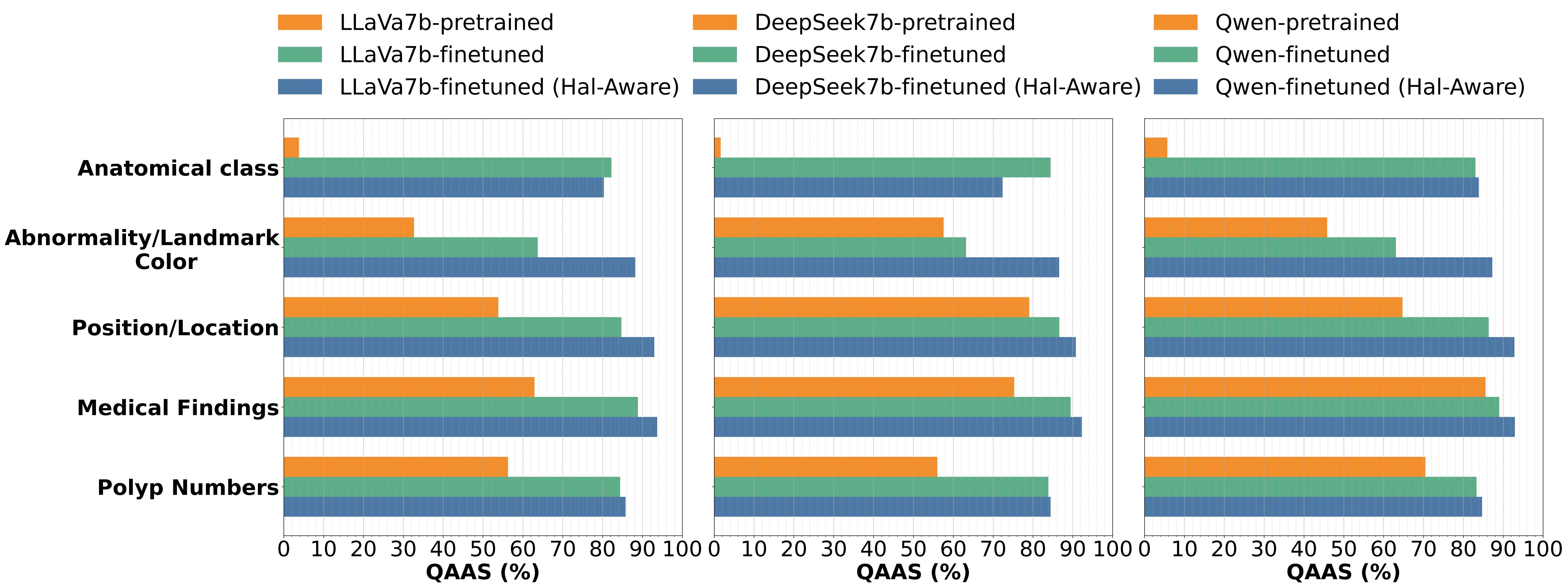}
    \caption{Comparison of VLM responses evaluated across different aspects}
    \label{fig:comparison}
    \vspace{-1em}
\end{figure}

There was a notable improvement across all metrics when finetuning VLMs with ground-truth texts. However, the hallucination-aware finetuning (finetuned\textsuperscript{H}) outperformed the standard finetuning, suggesting that training the model to detect and correct hallucinations leads to improved performance and produces more reliable, context-aware models. For instance, the LLaVA-1.6-7B finetuned\textsuperscript{H} achieved a QAAS score of $90.89\%$, compared to $83.07\%$ for its standard finetuned counterpart and only $50.89\%$ for pretrained version. Similarly, R-Sim improved from 1.36 (pretrained) to 3.71 (standard finetuning), with further improvement to 3.96 through hallucination-aware finetuning, demonstrating a more accurate semantic alignment with expert-generated descriptions. We could not finetune ChatGPT-4 Omni as it is not an open-source model. We also evaluated QAAS across different aspects as depicted in Fig. \ref{fig:comparison}, which shows that hallucination-aware training consistently outperforms standard training across most aspects.

\noindent\textit{\textbf{Expert Evaluation:} For LLaVa-1.6-7B, the average scores are $1.90$ (pretrained), $3.24$ (finetuned), and $3.36$ (hallucination-aware finetuned), with the latter scoring the highest. We also computed the rater's average coefficient of variation \cite{Wikipedia_CV} using the stand-out sample and found it to be $11.41\%$, indicating fair consistency. }

\subsubsection{ Visual Question Answering (VQA) Evaluation:}
Table \ref{tab:Results} also summarizes the performance of LLaVA-1.6-7B in VQA task. We focus on QAAS, as other metrics are mainly applicable only for description comparison. The pretrained models struggled to answer expert-designed questions effectively, achieving significantly lower QAAS, while finetuning substantially improved the performance.

\section{Discussion and Conclusion}

Here, we introduced Gut-VLM, a multimodal dataset for GI image analysis that includes hallucination-aware annotations to advance research on reliable and trustworthy VLMs. Our annotation process of using VLM-generated descriptive diagnosis reports, followed by expert corrections, not only reduces annotation costs by avoiding the need to hire experts for routine annotation from start but also enables the creation of a dataset with tags identifying potential hallucination patterns. We argue that by formulating VLM finetuning tasks as hallucination detection and correction rather than just diagnostic report generation, we can elicit reasoning in VLMs—similar to how engagement via error identification and correction enhances learning in humans compared to passive learning. 

This work also has some limitations. Budget constraints led to corrections being limited to responses from a single VLM (ChatGPT), potentially introducing bias toward its response structure. Additionally, our dataset offers sentence-level hallucination tags, which may limit granularity. We will expand the dataset by incorporating diverse VLM responses, annotating segment-level hallucination tags, and extending the dataset to include diverse demographics.

Future research could explore alternative hallucination detection and mitigation strategies, such as uncertainty estimation \cite{farquhar2024detecting}, reinforcement learning \cite{gunjal2024detecting}, architectural modifications \cite{llava1.5}, and feature fusion \cite{mmllversatile}. We also highlight the need for standardized benchmarks in the medical domain to accurately assess hallucinations and ensure reliable model evaluation, as well as ways to test the statistical significance of performance and its relevance to diagnosis.

%
%
\bibliographystyle{splncs04}
\bibliography{mybibliography}

\begin{thebibliography}{10}
\providecommand{\url}[1]{\texttt{#1}}
\providecommand{\urlprefix}{URL }
\providecommand{\doi}[1]{https://doi.org/#1}

\bibitem{achiam2023gpt}
Achiam, J., Adler, S., Agarwal, S., Ahmad, L., Akkaya, I., Aleman, F.L., Almeida, D., Altenschmidt, J., Altman, S., Anadkat, S., et~al.: {GPT-4} technical report. arXiv preprint arXiv:2303.08774  (2023)

\bibitem{arnold2020global}
Arnold, M., Abnet, C.C., Neale, R.E., Vignat, J., Giovannucci, E.L., McGlynn, K.A., Bray, F.: Global burden of 5 major types of gastrointestinal cancer. Gastroenterology  \textbf{159}(1),  335--349 (2020)

\bibitem{hallucinationmultimodallargelanguage}
Bai, Z., Wang, P., Xiao, T., He, T., Han, Z., Zhang, Z., Shou, M.Z.: Hallucination of multimodal large language models: A survey (2024), \url{https://arxiv.org/abs/2404.18930}

\bibitem{banerjee2005meteor}
Banerjee, S., Lavie, A.: {METEOR}: An automatic metric for {MT} evaluation with improved correlation with human judgments. In: Proceedings of the ACL workshop on intrinsic and extrinsic evaluation measures for machine translation and/or summarization. pp. 65--72 (2005)

\bibitem{hyperkvasir}
Borgli, H., Thambawita, V., Smedsrud, P.H., Hicks, S., Jha, D., Eskeland, S.L., Randel, K.R., Pogorelov, K., Lux, M., Nguyen, D.T.D., et~al.: Hyperkvasir, a comprehensive multi-class image and video dataset for gastrointestinal endoscopy. Scientific data  \textbf{7}(1), ~283 (2020)

\bibitem{hallucinationinmedical}
Chen, J., Yang, D., Wu, T., Jiang, Y., Hou, X., Li, M., Wang, S., Xiao, D., Li, K., Zhang, L.: Detecting and evaluating medical hallucinations in large vision language models (2024), \url{https://arxiv.org/abs/2406.10185}

\bibitem{AnamolyasObjectdet}
Chheda, T., Iyer, R., Koppaka, S., Kalbande, D.: Gastrointestinal tract anomaly detection from endoscopic videos using object detection approach. In: Bebis, G., Yin, Z., Kim, E., Bender, J., Subr, K., Kwon, B.C., Zhao, J., Kalkofen, D., Baciu, G. (eds.) Advances in Visual Computing. pp. 494--505. Springer International Publishing, Cham (2020)

\bibitem{deepseekai2025deepseekv3technicalreport}
DeepSeek-AI, Liu, A., Feng, B., Xue, B., Wang, B., et~al.: {DeepSeek-V3} technical report (2025), \url{https://arxiv.org/abs/2412.19437}

\bibitem{multimodalimage}
Du, H., Dong, Z., Wu, L., Li, Y., Liu, J., Luo, C., Zeng, X., Deng, Y., Cheng, D., Diao, W., Zhu, Y., Tao, X., Wang, J., Zhang, C., Yu, H.: A deep-learning based system using multi-modal data for diagnosing gastric neoplasms in real-time (with video). Gastric Cancer  \textbf{26}(2),  275--285 (Mar 2023)

\bibitem{farquhar2024detecting}
Farquhar, S., Kossen, J., Kuhn, L., Gal, Y.: Detecting hallucinations in large language models using semantic entropy. Nature  \textbf{630}(8017),  625--630 (2024)

\bibitem{kvasirvqa}
Gautam, S., Storås, A.M., Midoglu, C., Hicks, S.A., Thambawita, V., Halvorsen, P., Riegler, M.A.: {Kvasir-VQA}: A text-image pair gi tract dataset. In: Proceedings of the First International Workshop on Vision-Language Models for Biomedical Applications. p. 3–12. MM ’24, ACM (Oct 2024). \doi{10.1145/3689096.3689458}

\bibitem{gunjal2024detecting}
Gunjal, A., Yin, J., Bas, E.: Detecting and preventing hallucinations in large vision language models. In: Proceedings of the AAAI Conference on Artificial Intelligence (2024)

\bibitem{Medvqa}
Hicks, S., Stor{\aa}s, A.M., Halvorsen, P., de~Lange, T., Riegler, M., Thambawita, V.L.: Overview of {ImageCLEFmedical 2023} - medical visual question answering for gastrointestinal tract. In: Conference and Labs of the Evaluation Forum (2023)

\bibitem{hu2022lora}
Hu, E.J., Shen, Y., Wallis, P., Allen-Zhu, Z., Li, Y., Wang, S., Wang, L., Chen, W., et~al.: {LoRA}: Low-rank adaptation of large language models. ICLR  \textbf{1}(2), ~3 (2022)

\bibitem{CIEM}
Hu, H., Zhang, J., Zhao, M., Sun, Z.: {CIEM}: Contrastive instruction evaluation method for better instruction tuning (2023), \url{https://arxiv.org/abs/2309.02301}

\bibitem{kvasir-instrument}
Jha, D., Ali, S., Emanuelsen, K., Hicks, S.A., VajiraThambawita, Garcia-Ceja, E., Riegler, M.A., de~Lange, T., Schmidt, P.T., Johansen, H.D., Johansen, D., Halvorsen, P.: Kvasir-instrument: Diagnostic and therapeutic tool segmentation dataset in gastrointestinal endoscopy (2020), \url{https://arxiv.org/abs/2011.08065}

\bibitem{gastrovision}
Jha, D., Sharma, V., Dasu, N., Tomar, N.K., Hicks, S., Bhuyan, M.K., Das, P.K., Riegler, M.A., Halvorsen, P., Bagci, U., et~al.: Gastrovision: A multi-class endoscopy image dataset for computer aided gastrointestinal disease detection. In: Workshop on Machine Learning for Multimodal Healthcare Data. pp. 125--140. Springer (2023)

\bibitem{comthallucination}
Jiang, Y., Chen, J., Yang, D., Li, M., Wang, S., Wu, T., Li, K., Zhang, L.: {CoMT}: Chain-of-medical-thought reduces hallucination in medical report generation (2024), \url{https://arxiv.org/abs/2406.11451}

\bibitem{faithscore}
Jing, L., Li, R., Chen, Y., Du, X.: {FaithScore}: Fine-grained evaluations of hallucinations in large vision-language models (2024), \url{https://arxiv.org/abs/2311.01477}

\bibitem{keshavarz2024chatgpt}
Keshavarz, P., Bagherieh, S., Nabipoorashrafi, S.A., Chalian, H., Rahsepar, A.A., Kim, G.H.J., Hassani, C., Raman, S.S., Bedayat, A.: {ChatGPT} in radiology: A systematic review of performance, pitfalls, and future perspectives. Diagnostic and interventional imaging  (2024)

\bibitem{li2022deep}
Li, J., Hu, S., Shi, C., Dong, Z., Pan, J., Ai, Y., Liu, J., Zhou, W., Deng, Y., Li, Y., et~al.: A deep learning and natural language processing-based system for automatic identification and surveillance of high-risk patients undergoing upper endoscopy: a multicenter study. EClinicalMedicine  \textbf{53} (2022)

\bibitem{POPE}
Li, Y., Du, Y., Zhou, K., Wang, J., Zhao, W.X., Wen, J.R.: Evaluating object hallucination in large vision-language models (2023), \url{https://arxiv.org/abs/2305.10355}

\bibitem{lin2004rouge}
Lin, C.Y.: {ROUGE}: A package for automatic evaluation of summaries. In: Text summarization branches out. pp. 74--81 (2004)

\bibitem{lin2014microsoft}
Lin, T.Y., Maire, M., Belongie, S., Hays, J., Perona, P., Ramanan, D., Doll{\'a}r, P., Zitnick, C.L.: Microsoft coco: Common objects in context. In: Computer vision--ECCV 2014: 13th European conference, zurich, Switzerland, September 6-12, 2014, proceedings, part v 13. pp. 740--755. Springer (2014)

\bibitem{LRVInstruction}
Liu, F., Lin, K., Li, L., Wang, J., Yacoob, Y., Wang, L.: Mitigating hallucination in large multi-modal models via robust instruction tuning (2024), \url{https://arxiv.org/abs/2306.14565}

\bibitem{Gaive}
Liu, F., Lin, K., Li, L., Wang, J., Yacoob, Y., Wang, L.: Mitigating hallucination in large multi-modal models via robust instruction tuning (2024), \url{https://arxiv.org/abs/2306.14565}

\bibitem{llava1.5}
Liu, H., Li, C., Li, Y., Lee, Y.J.: Improved baselines with visual instruction tuning (2024), \url{https://arxiv.org/abs/2310.03744}

\bibitem{liu2023visual}
Liu, H., Li, C., Wu, Q., Lee, Y.J.: Visual instruction tuning. Advances in neural information processing systems  \textbf{36},  34892--34916 (2023)

\bibitem{lorraine2023diagnostic}
Lorraine-Francis, H., Newberry, E., Aziz, I.: Diagnostic yield of upper gastrointestinal endoscopy in patients attending a {UK} centre with symptoms compatible with {Rome IV} functional dyspepsia. Frontline Gastroenterology  \textbf{14}(4),  306--311 (2023)

\bibitem{marques2017image}
Marques, S., Bispo, M., Pimentel-Nunes, P., Chagas, C., Dinis-Ribeiro, M.: Image documentation in gastrointestinal endoscopy: review of recommendations. GE-Portuguese Journal of Gastroenterology  \textbf{24}(6),  269--274 (2017)

\bibitem{svmcnngastrodisease}
Melaku Bitew~Haile, Ayodeji Olalekan~Salau, B.E., Belay, A.J.: Detection and classification of gastrointestinal disease using convolutional neural network and svm. Cogent Engineering  \textbf{9}(1),  2084878 (2022)

\bibitem{metcalfe2017learning}
Metcalfe, J.: Learning from errors. Annual review of psychology  \textbf{68}(1),  465--489 (2017)

\bibitem{papineni2002bleu}
Papineni, K., Roukos, S., Ward, T., Zhu, W.J.: {BLEU}: a method for automatic evaluation of machine translation. In: Proceedings of the 40th annual meeting of the Association for Computational Linguistics. pp. 311--318 (2002)

\bibitem{kvasir}
Pogorelov, K., Randel, K.R., Griwodz, C., Eskeland, S.L., de~Lange, T., Johansen, D., Spampinato, C., Dang-Nguyen, D.T., Lux, M., Schmidt, P.T., et~al.: Kvasir: A multi-class image dataset for computer aided gastrointestinal disease detection. In: Proceedings of the 8th ACM on Multimedia Systems Conference. pp. 164--169 (2017)

\bibitem{ncdd}
Pokhrel, S., Bhandari, S., Ali, S., Lambrou, T., Nguyen, A., Shrestha, Y.R., Watson, A., Stoyanov, D., Gyawali, P., Bhattarai, B.: {NCDD}: Nearest centroid distance deficit for out-of-distribution detection in gastrointestinal vision (2024), \url{https://arxiv.org/abs/2412.01590}

\bibitem{ttaood}
Pokhrel, S., Bhandari, S., Vazquez, E., Lambrou, T., Gyawali, P., Bhattarai, B.: {TTA-OOD}: Test-time augmentation for improving out-of-distribution detection in gastrointestinal vision (2024), \url{https://arxiv.org/abs/2407.14024}

\bibitem{selfsupervisedood}
Quindós, A., Laiz, P., Vitrià, J., Seguí, S.: Self-supervised out-of-distribution detection in wireless capsule endoscopy images. Artificial Intelligence in Medicine  \textbf{143},  102606 (2023)

\bibitem{CHAIR}
Rohrbach, A., Hendricks, L.A., Burns, K., Darrell, T., Saenko, K.: Object hallucination in image captioning (2019), \url{https://arxiv.org/abs/1809.02156}

\bibitem{selivanov2023medical}
Selivanov, A., Rogov, O.Y., Chesakov, D., Shelmanov, A., Fedulova, I., Dylov, D.V.: Medical image captioning via generative pretrained transformers. Scientific Reports  \textbf{13}(1), ~4171 (2023)

\bibitem{giclassification}
Sharmila, V., Geetha, S.: Detection and classification of gi-tract anomalies from endoscopic images using deep learning. In: 2022 IEEE 19th India Council International Conference (INDICON). pp.~1--6 (Nov 2022)

\bibitem{shieh2024assessing}
Shieh, A., Tran, B., He, G., Kumar, M., Freed, J.A., Majety, P.: Assessing {ChatGPT 4.0’s} test performance and clinical diagnostic accuracy on {USMLE STEP 2 CK} and clinical case reports. Scientific Reports  \textbf{14}(1), ~9330 (2024)

\bibitem{MeidcalVLpretraining}
Shrestha, P., Amgain, S., Khanal, B., Linte, C.A., Bhattarai, B.: Medical vision language pretraining: A survey (2023), \url{https://arxiv.org/abs/2312.06224}

\bibitem{singhal2025toward}
Singhal, K., Tu, T., Gottweis, J., Sayres, R., Wulczyn, E., Amin, M., Hou, L., Clark, K., Pfohl, S.R., Cole-Lewis, H., et~al.: Toward expert-level medical question answering with large language models. Nature Medicine pp.~1--8 (2025)

\bibitem{gitractensemblestacking}
Sivari, E., Bostanci, E., Guzel, M.S., Acici, K., Asuroglu, T., Ercelebi~Ayyildiz, T.: A new approach for gastrointestinal tract findings detection and classification: Deep learning-based hybrid stacking ensemble models. Diagnostics  \textbf{13}(4) (2023)

\bibitem{kvasir-capsule}
Smedsrud, P.H., Thambawita, V., Hicks, S.A., Gjestang, H., Nedrejord, O.O., N{\ae}ss, E., Borgli, H., Jha, D., Berstad, T.J.D., Eskeland, S.L., Lux, M., Espeland, H., Petlund, A., Nguyen, D.T.D., Garcia-Ceja, E., Johansen, D., Schmidt, P.T., Toth, E., Hammer, H.L., de~Lange, T., Riegler, M.A., Halvorsen, P.: {Kvasir-Capsule}, a video capsule endoscopy dataset. Scientific Data  \textbf{8}(1), ~142 (May 2021)

\bibitem{mmllversatile}
Tong, S., Liu, Z., Zhai, Y., Ma, Y., LeCun, Y., Xie, S.: Eyes wide shut? exploring the visual shortcomings of multimodal llms (2024), \url{https://arxiv.org/abs/2401.06209}

\bibitem{Recaption}
Wang, L., He, J., Li, S., Liu, N., Lim, E.P.: Mitigating fine-grained hallucination by fine-tuning large vision-language models with caption rewrites (2023), \url{https://arxiv.org/abs/2312.01701}

\bibitem{Wikipedia_CV}
{Wikipedia contributors}: Coefficient of variation --- {Wikipedia}, the free encyclopedia (2025), \url{https://en.wikipedia.org/wiki/Coefficient_of_variation}, [Online; accessed 27-February-2025]

\bibitem{yang2024qwen2technicalreport}
Yang, A., Yang, B., Hui, B., Zheng, B., Yu, B., et~al.: Qwen2 technical report (2024), \url{https://arxiv.org/abs/2407.10671}

\bibitem{ye2024mplugowl3longimagesequenceunderstanding}
Ye, J., Xu, H., Liu, H., Hu, A., Yan, M., Qian, Q., Zhang, J., Huang, F., Zhou, J.: {mPLUG-Owl3}: Towards long image-sequence understanding in multi-modal large language models (2024), \url{https://arxiv.org/abs/2408.04840}

\bibitem{hallucidoctor}
Yu, Q., Li, J., Wei, L., Pang, L., Ye, W., Qin, B., Tang, S., Tian, Q., Zhuang, Y.: {HalluciDoctor}: Mitigating hallucinatory toxicity in visual instruction data (2024), \url{https://arxiv.org/abs/2311.13614}

\bibitem{eos}
Yue, Z., Zhang, L., Jin, Q.: Less is more: Mitigating multimodal hallucination from an eos decision perspective (2024), \url{https://arxiv.org/abs/2402.14545}

\bibitem{zebra_healthcare}
{Zebra Technologies}: Healthcare solutions (2024), \url{https://www.zebra.com/us/en/industry/healthcare.html}, accessed: 2024-02-23

\end{thebibliography}

\end{document}